\documentclass[conference]{IEEEtran}
\IEEEoverridecommandlockouts

\usepackage[T1]{fontenc}
\usepackage[style=ieee]{biblatex}

\usepackage{amsmath,amssymb,amsfonts}
\usepackage{algorithmic}
\usepackage{graphicx}
\usepackage{textcomp}
\usepackage{xcolor}
\usepackage{booktabs}

\usepackage{hyperref}
\usepackage{xfrac}
\usepackage{flushend}

\def\BibTeX{{\rm B\kern-.05em{\sc i\kern-.025em b}\kern-.08em
    T\kern-.1667em\lower.7ex\hbox{E}\kern-.125emX}}
\begin{document}

\title{Scaling Scaling Laws with Board Games}

\author{\IEEEauthorblockN{Andy L. Jones} \\
\IEEEauthorblockA{
London, United Kingdom \\
me@andyljones.com}}

\maketitle

\begin{abstract}
The largest experiments in machine learning now require resources far beyond the budget of all but a few institutions. Fortunately, it has recently been shown that the results of these huge experiments can often be extrapolated from the results of a sequence of far smaller, cheaper experiments. In this work, we show that not only can the extrapolation be done based on the size of the model, but on the size of the problem as well. By conducting a sequence of experiments using AlphaZero and Hex, we show that the performance achievable with a fixed amount of compute degrades predictably as the game gets larger and harder. Along with our main result, we further show that the test-time and train-time compute available to an agent can be traded off while maintaining performance.
\end{abstract}


\section{Introduction}
There is a concern that the state-of-the-art models studied by the most well-resourced organisations are growing too expensive for other researchers to keep pace \cite{strubell2019energy, stanford2020nrc, ukri2021transforming}. Fortunately, the recently-proposed paradigm of \emph{scaling laws} proposes a solution: that by studying the behaviour of a sequence of small, cheap models, researchers can extrapolate the behaviour of large, expensive models without having to explicitly train them.

In the past year, scaling laws have been established over a range of domains in machine learning \cite{hestness2017deep, rosenfeld2019constructive, kaplan2020scaling, henighan2020scaling, rosenfeld2020predictability, hernandez2021scaling}. These laws show that the performance of each model in a family can be well-characterised by a function some `size' property (like data or compute), and that the function behaves predictably over many orders of magnitude in model size.

So far however these works have only considered scaling the size of the model, leaving fixed the problem under consideration. Our principal contribution is to generalise this, scaling not only the model but the problem as well. In this work, we show that the behaviour of a model on a small problem instance predicts the behaviour of a model on a much larger problem instance. 

Our problem of choice is the board game Hex \cite{wiki2021hex}, a strategic board game whose complexity can be easily adjusted by changing the board size. Using AlphaZero \cite{silver2018general}, we train many different models on many different board sizes. Analysed together, the performance of these models reveals a \emph{compute frontier} that bounds the performance a model from our family in terms of the compute used to train it. These compute frontiers are exponential in the desired performance, and exponential again in the board size. 

Building on these results, we show that compute frontiers fitted at small board sizes are good predictors of the compute frontiers discovered at large board sizes. More, the error in the prediction drops exponentially as more small board sizes are added to the fit. 

Finally, while pursuing our main results we discovered an independently-interesting result: that for each extra order of magnitude of train-time compute, we can reduce test-time compute by a similar factor  while leaving performance unchanged.

We have published our code, models and data on GitHub\footnote{https://andyljones.com/boardlaw/}. 

\section{Background}

\subsection{Scaling Laws}
While the general idea of studying power laws in model size stretches back to at least the 1980s \cite{devore1989optimal}, it was the work of Hestness et al. \cite{hestness2017deep} that first brought the phenomenon to the attention of a contemporary audience. Their work showed that over a range of network architectures, the performance of a language model followed a power-law in the size of the dataset it was trained on.

Later, Rosenfeld et al. \cite{rosenfeld2019constructive} showed that the fit of the power law could be substantially improved by taking into account the size of the model, while Kaplan et al. \cite{kaplan2020scaling} further added the amount of compute spent training it. Then in Henighan et al. \cite{henighan2020scaling}, these laws were further shown to hold -- with varying coefficients -- over a range of generative modelling tasks, including video. Most recently Hernandez et al. \cite{hernandez2021scaling} have shown laws in fine-tuning, and Rosenfeld et al. \cite{rosenfeld2020predictability} in pruning. 

There has also been work on the theoretical underpinnings of these laws. Hutter \cite{hutter2021learning} is the most recent contribution in the area, and its introduction provides an exhaustive overview of prior work.

So far however, published work on scaling laws has exclusively addressed images and language. The forthcoming Hilton et al. \cite{hilton2021scalingrl} studies scaling laws in single-agent reinforcement learning, but ours is the first work on scaling laws in multi-agent reinforcement learning, and the first to scale the size of the problem as well as the size of the model. 

\subsection{AlphaZero}
AlphaZero \cite{silver2018general} is an algorithm for teaching a neural network to play a two-player zero-sum game entirely through self-play. At each step in the training process, AlphaZero augments the network-generated policy with a tree search. The augmented policy is stronger than the original policy on its own, and consequently self-play games between the augmented network and itself can be used as a source of experience to train the network. This amplification process \cite{christiano2019alphazero} progressively bootstraps the network from a random initialisation up to superhuman play, and - importantly - does so in a way that requires no extra human input.
 
\subsection{Hex}
\begin{figure}[tb]
\centerline{\includegraphics[width=.5\textwidth]{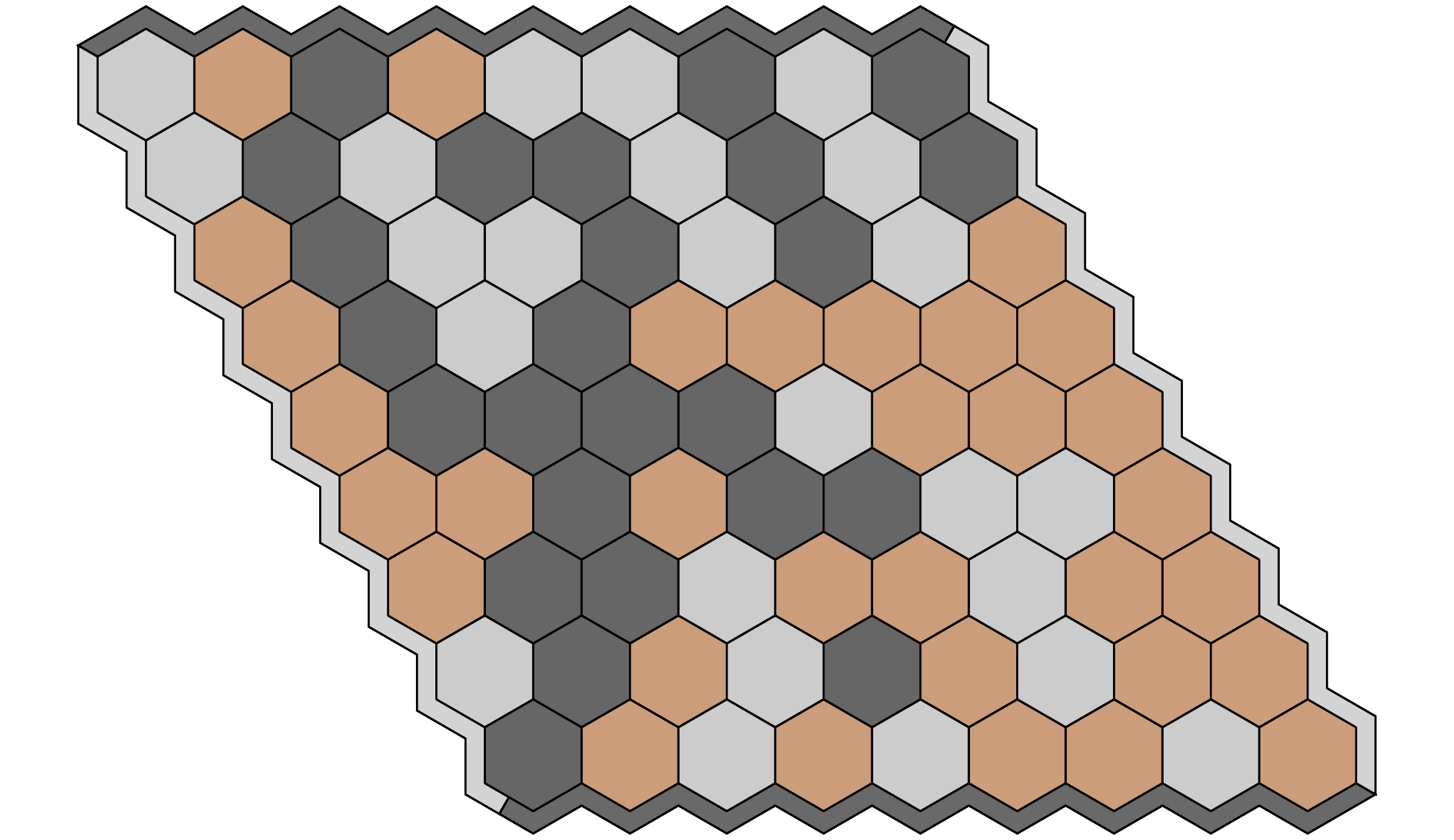}}
\caption{A Hex game on a $9 \times 9$ board, won by black with the path in the second column.}
\label{hex}
\end{figure}

Hex \cite{wiki2021hex} is a strategy game for two players. The players take turns placing tokens on a rhombic board, and the first player to connect their sides of the board is the winner (Fig \ref{hex}). First developed by Hein in 1942 \cite{hein1942vil}, Hex has enjoyed niche popularity throughout its life \cite{hayward2019hex}. 

Despite the simplicity of its rule set, Hex is considered to be a complex strategic game \cite{seymour2021hex}. In fact, despite sustained attention from games researchers \cite{huang2013mohex, young2016neurohex, anthony2017thinking}, computers only surpassed human-level play at large board sizes in 2020 \cite{cazenave2020polygames}.

We chose Hex as the focus of our work because it is easy to vary the size and complexity of the game, and because it is easy to implement as a fast, parallelized GPU kernel. More popular games such as Chess, Go and Shogi have all accumulated minor rules - such as castling,  \emph{kō} or \emph{nifu} - that make for dramatically more complex and bug-prone implementations \cite{chessprog2021}. 

One further simplification we make is that while human games of Hex are typically played with the `pie rule' as a way to nullify first-mover advantage, in our implementation we omit it. Instead, all evaluation matches are played as a pair, with each agent playing black in one match and white in the other. 

\subsection{Ratings and Elo}
\label{background-elo}
Unlike in regular reinforcement learning where performance (reward) can be measured easily, the performance of an agent in a multiplayer game depends  on who the opponent is. As such, any rating system needs to take account of not only the player but also their opponent. 

\begin{figure}[tb]
\centerline{\includegraphics[width=.5\textwidth]{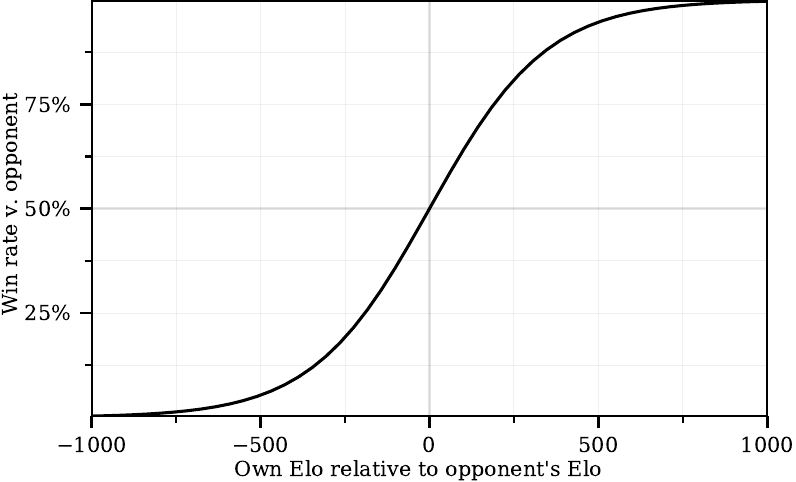}}
\caption{The Elo ratings of two players predict the outcome of a match between them, with the player with the higher Elo being more likely to win.}
\label{elos}
\end{figure}

In human chess tournaments, the solution is the Elo system \cite{elo1978rating}. The Elo system assigns each player a numerical ranking - their \emph{Elo} - in such a way that that the chance of one player winning over another can be calculated from the difference between the two player's Elos (Fig \ref{elos}). Stronger players come out of this system with high Elos; weak players with low Elos. 

The central limitation of the Elo system is that it assumes \emph{transitivity}. This is not necessarily the case, and in fact there are games - such as rock-paper-scissors - where the Elos assigned to each player are entirely uninformative \cite{balduzzi2018reevaluating, rowland2020multiagent, czarnecki2020real}.

Elo is also a \emph{relative} rating system, meaning that any set of Elo ratings can be shifted up or down by a constant offset without affecting their predictive ability. Fortunately, on our board sizes there is an excellent choice of constant offset: fixing perfect play to zero Elo. MoHex \cite{MoHex, pawlewicz2014stronger, huang2013mohex, pawlewicz2013scalable} is an algorithmic agent that can play perfectly on board sizes up to $9 \times 9$, and we fix its play to zero for all Elo ratings reported herein. 

While Elo is the best known rating system of its type, there are other more modern variations such as Glicko \cite{glickman1995glicko} and TrueSkill \cite{minka2007trueskill}. These variations are all more complex however, and the additional complexities would not improve the analyses carried out in this work. 

\section{Methods}
We developed a fast, low-resource AlphaZero implementation (documented in Appendix \ref{alphazero}) and used it to train many different models on many different board sizes. We then evaluated the trained models against perfect play in order to come up with compute frontiers at each board size. Finally, we fitted a simple curve to these frontiers, to show that the relationship is consistent across board sizes.  

\subsection{AlphaZero}

\begin{figure}[tb]
\centerline{\includegraphics[width=.49\textwidth]{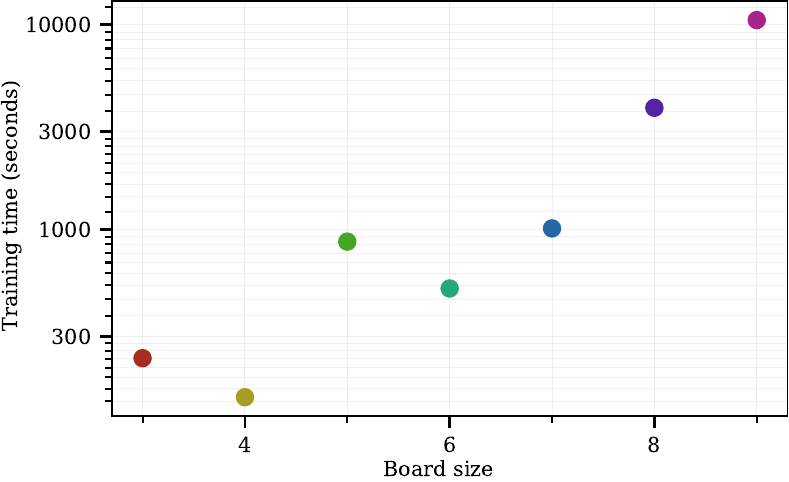}}
\caption{The time taken to train an agent to -50 Elo (ie, almost equal to perfect play) is roughly exponential in boardsize, with the fastest agent on a $9\times 9$ board taking  about 3 hours.}
\label{runtimes}
\end{figure}

Our implementation of AlphaZero can train an agent to perfect play in time approximately exponential in board size (Fig \ref{runtimes}). In particular, perfect play on a $9 \times 9$ board takes a little under 3 hours on a single RTX 2080 Ti. We have not been able to find good baselines for the performance of our implementation -- the only other $9 \times 9$ AlphaZero Hex implementation we know of is Polygames' \cite{cazenave2020polygames}, and training time figures have not been made available for it. 

\subsection{Models}
We used AlphaZero to train $\approx 200$ different models over a range of hyperparameters. Most hyperparameters were held constant across runs and are documented in Table \ref{hyperparams}, while a few - principally the network architectures and run duration - varied with the board size, and are documented in Table \ref{boardsize}.

\begin{table}[t]
\centering
\caption{Hyperparameters}
\label{hyperparams}
\begin{tabular}{ll}
\toprule
Number of envs        &              32k \\
Batch size            &              32k \\
Buffer size           &       2m samples \\
Learning rate         &             1e-3 \\
MCTS node count       &               64 \\
MCTS $c_\text{puct}$  &  $\sfrac{1}{16}$ \\
MCTS noise $\epsilon$ &   $\sfrac{1}{4}$ \\
\bottomrule
\end{tabular}
\end{table}

\begin{table}
\centering
\caption{Board size-dependent hyperparameter limits}
\label{boardsize}
\begin{tabular}{rrrll}
\toprule
 Board Size &  Neurons &  Layers & Samples & Compute \\
\midrule
          3 &        2 &       4 &   4E+08 &   1E+12 \\
          4 &       16 &       4 &   2E+08 &   1E+13 \\
          5 &       16 &       8 &   3E+08 &   3E+13 \\
          6 &      128 &       8 &   6E+08 &   4E+14 \\
          7 &      512 &       8 &   1E+09 &   1E+16 \\
          8 &      512 &       8 &   1E+09 &   3E+16 \\
          9 &     1024 &       8 &   2E+09 &   1E+17 \\
\bottomrule
\end{tabular}
\end{table}

The independent variables for our analysis are board size and compute. Varying board size is simple, but there are many ways to vary the amount of compute involved in training a model. We chose to explore three axes of compute variation: the depth of the network, the width of the network, and the length of the training run. Specifically,

\subsubsection{Board size} Board sizes ranged from 3 to 9. The smallest board used was $3 \times 3$, as this is the smallest `interesting' board size. The largest board used was $9 \times 9$, as this was the largest board MoHex can achieve perfect play on.

\subsubsection{Agent architecture} Agent architectures ranged in powers of 2 from 1 layer of 1 neuron through to 8 layers of 1024 neurons. The maximum agent size for each board size was determined during preliminary work, and is listed in Table \ref{boardsize}.

\subsubsection{Run length} Training runs were terminated when they hit a certain number of samples or a certain number of FLOPS-seconds. These limits were also determined during preliminary work, and are listed in Table \ref{boardsize}. 

\subsubsection{Snapshots} Snapshots were taken from the training run on a schedule exponential in compute. The schedule was chosen so that a training run hitting the compute limit would have 21 snapshots taken. Overall, we took 2,800 snapshots in total.

\subsection{Evaluation}
We evaluated the agents by playing each agent against each other agent for 1024 matches, with each agent playing black for 512 of those matches and white for the other 512. We chose this number of matches based on hardware,  time constraints, and the number of pairings that needed to be evaluated. We then used the outcomes from the matches to calculate an Elo rating for each agent.

Playing 1,024 matches between each pair of snapshots means playing 700m matches overall. To accelerate the evaluation, we took groups of 64 agents and played all 2m matches between them in parallel, batching the inferences for evaluation on the GPU. By fully saturating the GPU, we found we could play about 1k evaluation matches/GPU/second. 

\begin{figure}[tb]
\centerline{\includegraphics[width=.49\textwidth]{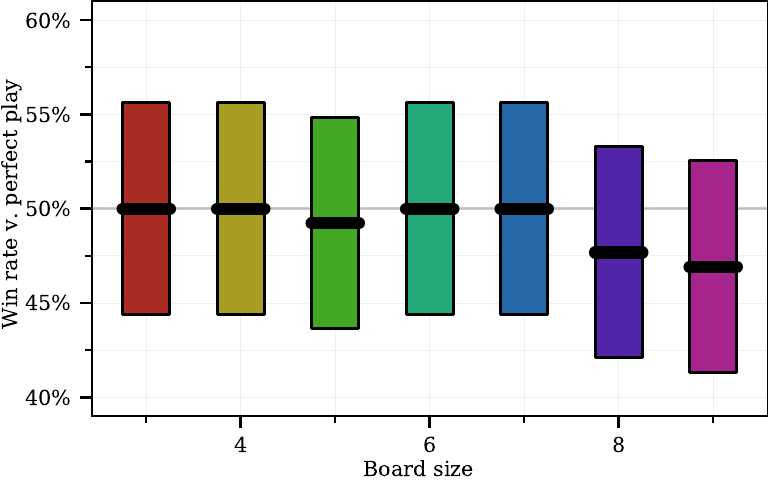}}
\caption{Our best AlphaZero agents are on par with MoHex's perfect play. Shown are the 90\% credible intervals on the best agents' win rate against MoHex after 128 games, assuming a $\text{Beta}(1, 1)$ prior.}
\label{calibration}
\end{figure}

While the matches between AlphaZero agents can establish the relative ratings, to fix the offset we also played the top-ranking agents against MoHex. The top-ranking agents reliably draw MoHex (Fig. \ref{calibration}), showing they are on par with perfect play.

\subsubsection{Hyperparameters} The same search hyperparameters were used for evaluation as were used during training, as listed in Table \ref{hyperparams}.

\subsection{Hardware} 
Each training run was conducted on a single RTX 2080 Ti, with many runs being carried out in parallel on machines rented from \href{vast.ai}{vast.ai}. In all, about 500 GPU-hours were used for training.

Evaluation matches meanwhile were carried out on two in-house RTX 2080 Tis, taking about 100 GPU-hours in all. 

\subsection{Curve fitting}
Having trained and evaluated the agents, the final step is to fit a functional form to the frontiers. The frontiers give the maximum performance attained for each quantity of compute at each board size, and can be roughly described as sequence of parallel plateaus, leading up into a set of parallel inclines, leading out onto a second plateau at zero Elo.

We explored several formalisations of this pattern (Appendix \ref{curve}) before settling on a five-parameter change-point model:

\begin{align*}
    \text{plateau} &= m^\text{plateau}_\text{boardsize} \cdot \text{boardsize} + c^\text{plateau} \\ 
    \text{incline} &= m^\text{incline}_\text{boardsize} \cdot \text{boardsize} + m^\text{incline}_\text{flops} \cdot \log \text{flop} + c^\text{incline} \\
    \text{elo} &= \text{incline}.\text{clamp}(\text{plateau}, 0)
\end{align*}

The first equation gives the lower set of parallel plateaus, the second the parallel inclines, and the third combines them. We fit the model with L-BFGS.

\section{Results}

\subsection{Frontier parameters}
\begin{figure}[tb]
\centerline{\includegraphics[width=.49\textwidth]{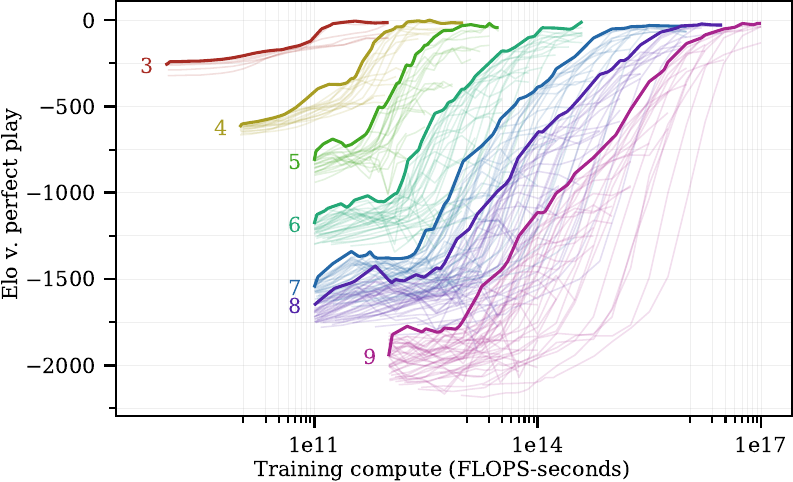}}
\caption{Each training run (each faint line) of each differently-sized agent follows a sigmoid, starting at random play and progressing up to some plateau. The frontiers (dark lines) formed by taking a maximum across training runs have a similar form across board sizes (colors).}
\label{flops_curves}
\end{figure}

During training, the performance of each agent describes a rough sigmoid in terms of compute spent (Fig. \ref{flops_curves}). Taking the maximum across agents at each level of compute gives the compute frontier, to which we fit our change-point model. 

\begin{figure}[tb]
\centerline{\includegraphics[width=.49\textwidth]{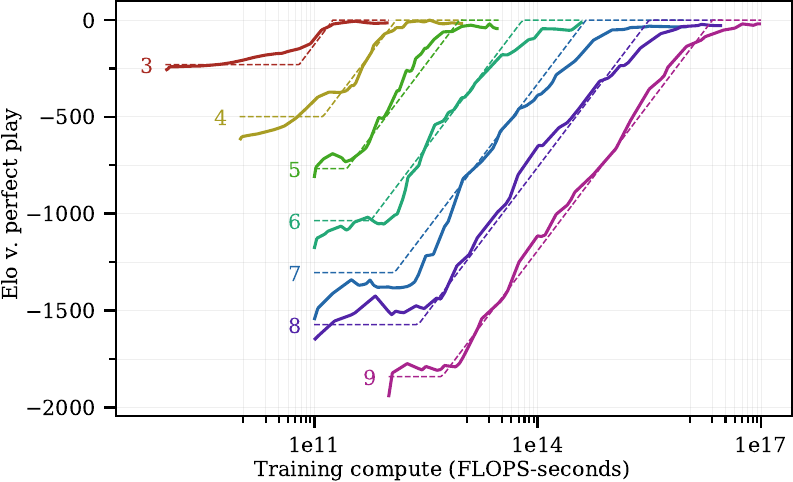}}
\caption{The compute-performance frontier follows the same sigmoid for each board size 3 through 9, just scaled and shifted. The dotted lines give the fitted curves.}
\label{frontiers}
\end{figure}

\begin{table}[t]
\centering
\caption{Fitted Frontier Parameters}
\label{parameters}
\begin{tabular}{llll}
\toprule
{} & $m_\text{flops}$ & $m_\text{boardsize}$ &    $c$ \\
\midrule
plateau &                  &                 -270 &    570 \\
incline &              510 &                 -430 &  -4400 \\
\bottomrule
\end{tabular}
\end{table}

The fitted frontiers are shown in Fig. \ref{frontiers}, and the parameters of those fits in Table \ref{parameters}. These parameters are easier to understand in terms of derived quantities:

\subsubsection{Slope} The slope of the incline is $500$ Elo per order of magnitude increase in compute. A more memorable interpretation is that if you are in the linearly-increasing regime, then you will need about $2 \times$ as much compute as your opponent to beat them $\sfrac{2}{3}$ of the time. 

\subsubsection{Perfect play} The minimum compute needed for perfect play increases $7\times$ for each increment in board size.

\subsubsection{Takeoff} The minimum training compute needed to see any improvement over random play increases by $4\times$ for each increment of board size.

\subsubsection{Random play} Finally, the distance between random play and perfect play increases by $500$ Elo for each increment of board size. Unlike the other quantities mentioned previously, the distance between random and perfect play is a property of the game itself rather than of the agent.

\subsection{Predictive errors}
While the model in the previous section was fitted across all board sizes simultaneously, we can alternatively ask: if we fit the model on data up to some small board size, how well does the fit predict the data from higher, unseen board sizes?

\begin{figure}[tb]
\centerline{\includegraphics[width=.49\textwidth]{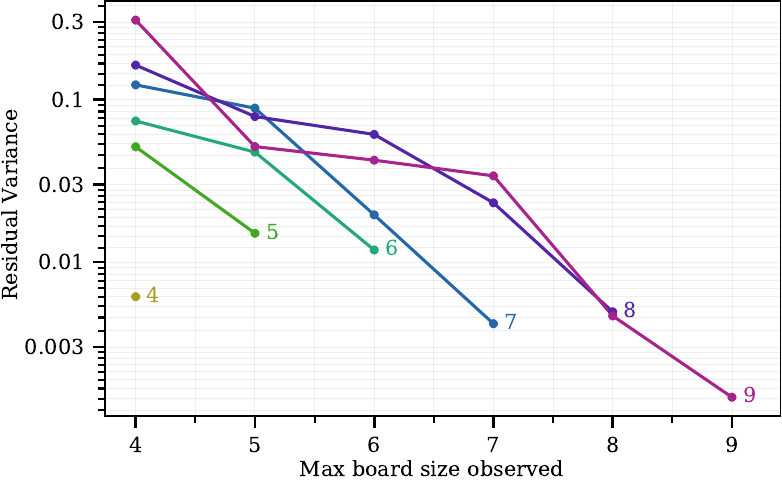}}
\caption{The error in the prediction decays exponentially as more boards are used. Each line gives the errors in the prediction for the frontier of a specific board size. }
\label{resid_var}
\end{figure}

As can be seen in Fig. \ref{resid_var}, the frontiers found at smaller board sizes accurately predict the frontiers that will be found at larger board sizes. The error in the predicted frontier (as measured by the residual variance) starts small and decays exponentially as more small boards are added to the fit. 

\subsection{Train-test trade-off}
While developing main results discussed above, a small unit of extra work was suggested towards an independently interesting result\footnote{Thanks and credit to Jared Kaplan for suggesting this.}.

\begin{figure}
\centerline{\includegraphics[width=.49\textwidth]{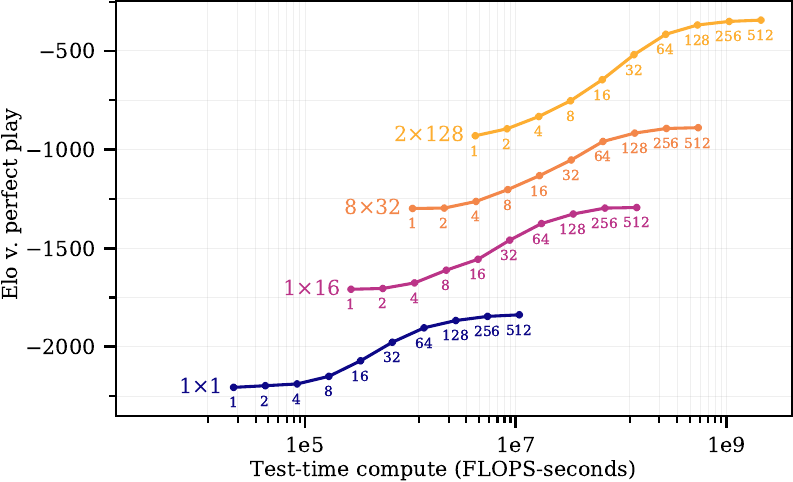}}
\caption{A selection of snapshots trained on a $9 \times 9$ board, evaluated with varying test-time tree sizes. These curves show that the performance of a specific snapshot is sigmoid in the test-time compute budget. The lines are labelled with the architecture of the snapshot, in the format $\text{depth} \times \text{width}$. Each point on the line is the Elo of that snapshot evaluated with a different tree size, spaced logarithmically between 1 node and 512 nodes.}
\label{test}
\end{figure}

So far we have focused on the compute budget during training, but another pertinent budget is the compute spent during evaluation. All the results discussed previously have used a tree search of size 64 during evaluation, the same as used during training. But there is no reason that the train-time search and test-time search have to be the same size, and so by varying the size of the test-time compute budget we can see in Fig. \ref{test} that larger tree searches at test time can substantially improve the performance of an agent.

\begin{figure}
\centerline{\includegraphics[width=.49\textwidth]{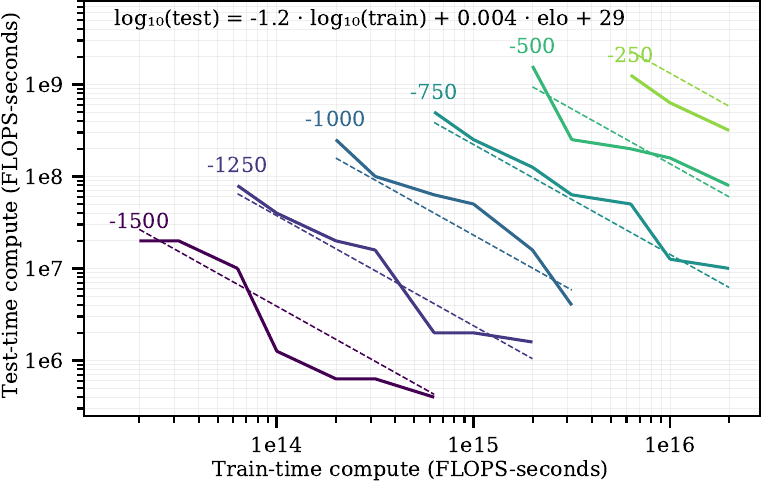}}
\caption{The trade-off between train-time compute and test-time compute. Each dotted line gives the minimum train-test compute required for a certain Elo on a $9 \times 9$ board}
\label{train_test}
\end{figure}

Knowing now that compute can be spent in two places, at train time and test time, the immediate question is: how do these two budgets trade off? This is illustrated in Fig. \ref{train_test}, which shows that the trade-off is linear in log-compute: for each additional $10\times$ of train-time compute, about $15\times$ of test-time compute can be eliminated, down to a floor of a single-node tree search.

\section{Discussion}
Our central, concrete result is that when we train AlphaZero to play Hex, the compute required can be calculated directly from the board size and the desired performance. We have also shown that compute during training and compute at test time can be traded off according to simple relationship. These results illuminate several intriguing phenomena.

First, the way in which performance scales with compute is that an agent with twice as much compute as its opponent can win roughly $\sfrac{2}{3}$ of the time. This behaviour is strikingly similar to that of a toy model where each player chooses as many random numbers as they have compute, and the player with the highest number wins\footnote{Thanks and credit to Paul Christiano for making us aware of this.}. In this toy model, doubling your compute doubles how many random numbers you draw, and the probability that you possess the largest number is $\sfrac{2}{3}$. This suggests that the complex game play of Hex might actually reduce to each agent having a `pool' of strategies proportional to its compute, and whoever picks the better strategy wins. While on the basis of the evidence presented herein we can only consider this to be serendipity, we are keen to see whether the same behaviour holds in other games.

Second, both the relation of performance to board size and the relation of performance to compute are smooth. Before embarking on this project, a key unknown was whether performance would show any `spikes' with regards to compute or board size. A spike with regards to compute might indicate the model had achieved some key insight, while a spike with regards to board size might indicate a minimum complexity past which key insights are available for the model to discover. As is however, models' performance changes smoothly and predictably with both increased compute and increased complexity. However, this could plausibly be a property unique to Hex and it's simple rule set, and we would again be keen to see whether the same behaviour holds in other games. 

Finally, the simple relationship between compute at train time and compute at test time was originally surprising to us. Our intuition was that test-time compute is much `cheaper' than train-time compute, and so we were surprised that one could easily substitute for the other. On reflection however, we believe the key distinction is that an optimization at test-time needs only optimise over one sample, while train-time compute meanwhile must optimise over the entire distribution of samples.

In all, these results demonstrate how a relationship between compute and performance identified in small, cheap problems carries directly over to problems sizes that are orders of magnitude more expensive to explore. If this phenomenon proves to be general, it opens the way for researchers to contribute to the understanding of problems far larger than the ones they themselves are able to directly study.

\section*{Acknowledgements}
This work was funded by \href{http://survivalandflourishing.org/}{Survival \& Flourishing}. This work has also benefited greatly from the advice of many friends and colleagues. In particular, we wish to acknowledge the invaluable input of Jared Kaplan, Jan Leike, Paul Christiano, Danny Hernandez, Jacob Hilton, Matthew Rahtz, Marc Lanctot, Max O. Smith, Ryan Hayward, Paul Lu, Adam Gleave, Asya Bergal, Mario Lezcano Casado, Ben Wang, Jeremy Salwen, Clemens Winter, and Ella Guest.


\printbibliography

@article{strubell2019energy,
  title={{Energy and policy considerations for deep learning in NLP}},
  author={Strubell, Emma and Ganesh, Ananya and McCallum, Andrew},
  journal={arXiv preprint arXiv:1906.02243},
  year={2019}
}

@article{stanford2020nrc,
  title={{National Research Cloud Call To Action}},
  author={{NRC Letter Signatories}},
  year="2020",
  url="https://hai.stanford.edu/national-research-cloud-joint-letter",
}

@article{ukri2021transforming,
    title="{Transforming our world with AI}",
    author={{UK Research and Innovation}},
    year="2021"
}

@article{hestness2017deep,
  title={Deep learning scaling is predictable, empirically},
  author={Hestness, Joel and Narang, Sharan and Ardalani, Newsha and Diamos, Gregory and Jun, Heewoo and Kianinejad, Hassan and Patwary, Md and Ali, Mostofa and Yang, Yang and Zhou, Yanqi},
  journal={arXiv preprint arXiv:1712.00409},
  year={2017}
}

@article{rosenfeld2019constructive,
  title={A constructive prediction of the generalization error across scales},
  author={Rosenfeld, Jonathan S and Rosenfeld, Amir and Belinkov, Yonatan and Shavit, Nir},
  journal={arXiv preprint arXiv:1909.12673},
  year={2019}
}

@article{kaplan2020scaling,
  title={Scaling laws for neural language models},
  author={Kaplan, Jared and McCandlish, Sam and Henighan, Tom and Brown, Tom B and Chess, Benjamin and Child, Rewon and Gray, Scott and Radford, Alec and Wu, Jeffrey and Amodei, Dario},
  journal={arXiv preprint arXiv:2001.08361},
  year={2020}
}

@article{henighan2020scaling,
  title={Scaling laws for autoregressive generative modeling},
  author={Henighan, Tom and Kaplan, Jared and Katz, Mor and Chen, Mark and Hesse, Christopher and Jackson, Jacob and Jun, Heewoo and Brown, Tom B and Dhariwal, Prafulla and Gray, Scott and others},
  journal={arXiv preprint arXiv:2010.14701},
  year={2020}
}

@article{hernandez2021scaling,
  title={Scaling Laws for Transfer},
  author={Hernandez, Danny and Kaplan, Jared and Henighan, Tom and McCandlish, Sam},
  journal={arXiv preprint arXiv:2102.01293},
  year={2021}
}

@article{rosenfeld2020predictability,
  title={On the Predictability of Pruning Across Scales},
  author={Rosenfeld, Jonathan S and Frankle, Jonathan and Carbin, Michael and Shavit, Nir},
  journal={arXiv preprint arXiv:2006.10621},
  year={2020}
}

@unpublished{hilton2021scalingrl,
    title={Scaling Laws for Reinforcement Learning},
    author={Hilton, Jacob and Tang, Jie},
    note={In preparation}
}

@article{devore1989optimal,
  title={Optimal nonlinear approximation},
  author={DeVore, Ronald A and Howard, Ralph and Micchelli, Charles},
  journal={Manuscripta mathematica},
  volume={63},
  number={4},
  pages={469--478},
  year={1989},
  publisher={Springer}
}

@article{hutter2021learning,
  title={Learning Curve Theory},
  author={Hutter, Marcus},
  journal={arXiv preprint arXiv:2102.04074},
  year={2021}
}

@article{silver2018general,
  title={A general reinforcement learning algorithm that masters chess, shogi, and Go through self-play},
  author={Silver, David and Hubert, Thomas and Schrittwieser, Julian and Antonoglou, Ioannis and Lai, Matthew and Guez, Arthur and Lanctot, Marc and Sifre, Laurent and Kumaran, Dharshan and Graepel, Thore and others},
  journal={Science},
  volume={362},
  number={6419},
  pages={1140--1144},
  year={2018},
  publisher={American Association for the Advancement of Science}
}

@article{christiano2019alphazero,
    title = "{AlphaGo Zero and capability amplification}",
    author = {Christiano, Paul},
    year = "2019",
    url = "https://www.lesswrong.com/posts/HA3oArypzNANvXC38/alphago-zero-and-capability-amplification",
}

@article{wiki2021hex,
    author = "{Wikipedia contributors}",
    title = "Hex --- {Wikipedia}{,} The Free Encyclopedia",
    year = "2020",
    url = "https://en.wikipedia.org/w/index.php?title=Hex\&oldid=996842461",
}

@article{seymour2021hex,
    author = {Seymour, Matthew},
    title = "Hex: A Strategy Guide",
    year = "2020",
    url = "http://www.mseymour.ca/hex_book/hexstrat.html",
}

@article{anthony2017thinking,
  title={Thinking fast and slow with deep learning and tree search},
  author={Anthony, Thomas and Tian, Zheng and Barber, David},
  journal={arXiv preprint arXiv:1705.08439},
  year={2017}
}

@incollection{young2016neurohex,
  title={Neurohex: A deep q-learning hex agent},
  author={Young, Kenny and Vasan, Gautham and Hayward, Ryan},
  booktitle={Computer Games},
  pages={3--18},
  year={2016},
  publisher={Springer}
}

@article{hein1942vil,
  title={Vil de laere Polygon?},
  author={Hein, Piet},
  journal={Politiken},
  year={1942},
  volume={December 26}
}

@book{hayward2019hex,
  title="{Hex: The Full Story}",
  author={Hayward, Ryan B and Toft, Bjarne},
  year={2019},
  publisher={CRC Press}
}

@article{chessprog2021,
    author = "{ChessProgramming contributors}",
    title = "Engine Testing",
    year = "2020",
    url = "https://www.chessprogramming.org/Engine\_Testing",
}

@book{elo1978rating,
  title={The rating of chessplayers, past and present},
  author={Elo, Arpad E},
  year={1978},
  publisher={Arco Pub.}
}

@article{glickman1995glicko,
  title={The glicko system},
  author={Glickman, Mark E},
  journal={Boston University},
  volume={16},
  pages={16--17},
  year={1995}
}

@article{minka2007trueskill,
  title={Trueskill tm: a Bayesian skill rating system},
  author={Minka, T and Graepel, T and Herbrich, R},
  journal={Advances in neural information processing systems},
  year={2007}
}

@article{balduzzi2018reevaluating,
      title={Re-evaluating Evaluation}, 
      author={David Balduzzi and Karl Tuyls and Julien Perolat and Thore Graepel},
      year={2018},
      eprint={1806.02643},
      archivePrefix={arXiv},
      primaryClass={cs.LG}
}

@article{rowland2020multiagent,
      title={Multiagent Evaluation under Incomplete Information}, 
      author={Mark Rowland and Shayegan Omidshafiei and Karl Tuyls and Julien Perolat and Michal Valko and Georgios Piliouras and Remi Munos},
      year={2020},
      eprint={1909.09849},
      archivePrefix={arXiv},
      primaryClass={cs.MA}
}

@article{czarnecki2020real,
      title={Real World Games Look Like Spinning Tops}, 
      author={Wojciech Marian Czarnecki and Gauthier Gidel and Brendan Tracey and Karl Tuyls and Shayegan Omidshafiei and David Balduzzi and Max Jaderberg},
      year={2020},
      eprint={2004.09468},
      archivePrefix={arXiv},
      primaryClass={cs.LG}
}

@article{MoHex,
  author = {{Henderson, Philip and Arneson, Broderick and Pawlewicz, Jakub and Huang, Aja and Young, Kenny and Gao, Chao}},
  title = {MoHex},
  url = {https://github.com/cgao3/benzene-vanilla-cmake},
  version = {d450c01},
  date = {2020-02-25},
}

@article{pawlewicz2014stronger,
  title={Stronger virtual connections in hex},
  author={Pawlewicz, Jakub and Hayward, Ryan and Henderson, Philip and Arneson, Broderick},
  journal={IEEE Transactions on Computational Intelligence and AI in Games},
  volume={7},
  number={2},
  pages={156--166},
  year={2014},
  publisher={IEEE}
}

@inproceedings{huang2013mohex,
  title={MoHex 2.0: a pattern-based MCTS Hex player},
  author={Huang, Shih-Chieh and Arneson, Broderick and Hayward, Ryan B and M{\"u}ller, Martin and Pawlewicz, Jakub},
  booktitle={International Conference on Computers and Games},
  pages={60--71},
  year={2013},
  organization={Springer}
}

@inproceedings{pawlewicz2013scalable,
  title={Scalable parallel DFPN search},
  author={Pawlewicz, Jakub and Hayward, Ryan B},
  booktitle={International Conference on Computers and Games},
  pages={138--150},
  year={2013},
  organization={Springer}
}

@article{lanctot2020openspiel,
      title={OpenSpiel: A Framework for Reinforcement Learning in Games}, 
      author={OpenSpiel Contributors},
      year={2020},
      eprint={1908.09453},
      archivePrefix={arXiv},
      primaryClass={cs.LG}
}

@article{tian2019elf,
      title={ELF OpenGo: An Analysis and Open Reimplementation of AlphaZero}, 
      author={Yuandong Tian and Jerry Ma and Qucheng Gong and Shubho Sengupta and Zhuoyuan Chen and James Pinkerton and C. Lawrence Zitnick},
      year={2019},
      eprint={1902.04522},
      archivePrefix={arXiv},
      primaryClass={cs.AI}
}

@article{wu2020accelerating,
      title={Accelerating Self-Play Learning in Go}, 
      author={David J. Wu},
      year={2020},
      eprint={1902.10565},
      archivePrefix={arXiv},
      primaryClass={cs.LG}
}

@article{cazenave2020polygames,
      title={Polygames: Improved Zero Learning}, 
      author={Tristan Cazenave and Yen-Chi Chen and Guan-Wei Chen and Shi-Yu Chen and Xian-Dong Chiu and Julien Dehos and Maria Elsa and Qucheng Gong and Hengyuan Hu and Vasil Khalidov and Cheng-Ling Li and Hsin-I Lin and Yu-Jin Lin and Xavier Martinet and Vegard Mella and Jeremy Rapin and Baptiste Roziere and Gabriel Synnaeve and Fabien Teytaud and Olivier Teytaud and Shi-Cheng Ye and Yi-Jun Ye and Shi-Jim Yen and Sergey Zagoruyko},
      year={2020},
      eprint={2001.09832},
      archivePrefix={arXiv},
      primaryClass={cs.LG}
}

@article{dalton2020accelerating,
      title={Accelerating Reinforcement Learning through GPU Atari Emulation}, 
      author={Steven Dalton and Iuri Frosio and Michael Garland},
      year={2020},
      eprint={1907.08467},
      archivePrefix={arXiv},
      primaryClass={cs.LG}
}

@article{megastep,
  author = {{Andy L Jones}},
  title = {megastep},
  url = {https://andyljones.com/megastep},
  version = {0.1},
  date = {2020-07-07},
}

@article{grill2020montecarlo,
      title={Monte-Carlo Tree Search as Regularized Policy Optimization}, 
      author={Jean-Bastien Grill and Florent Altché and Yunhao Tang and Thomas Hubert and Michal Valko and Ioannis Antonoglou and Rémi Munos},
      year={2020},
      eprint={2007.12509},
      archivePrefix={arXiv},
      primaryClass={cs.LG}
}

@article{stooke2019accelerated,
      title={Accelerated Methods for Deep Reinforcement Learning}, 
      author={Adam Stooke and Pieter Abbeel},
      year={2019},
      eprint={1803.02811},
      archivePrefix={arXiv},
      primaryClass={cs.LG}
}

@article{mccandlish2018empirical,
      title={An Empirical Model of Large-Batch Training}, 
      author={Sam McCandlish and Jared Kaplan and Dario Amodei and OpenAI Dota Team},
      year={2018},
      eprint={1812.06162},
      archivePrefix={arXiv},
      primaryClass={cs.LG}
}

@article{openai2019dota,
      title={Dota 2 with Large Scale Deep Reinforcement Learning}, 
      author={OpenAI and : and Christopher Berner and Greg Brockman and Brooke Chan and Vicki Cheung and Przemysław Dębiak and Christy Dennison and David Farhi and Quirin Fischer and Shariq Hashme and Chris Hesse and Rafal Józefowicz and Scott Gray and Catherine Olsson and Jakub Pachocki and Michael Petrov and Henrique P. d. O. Pinto and Jonathan Raiman and Tim Salimans and Jeremy Schlatter and Jonas Schneider and Szymon Sidor and Ilya Sutskever and Jie Tang and Filip Wolski and Susan Zhang},
      year={2019},
      eprint={1912.06680},
      archivePrefix={arXiv},
      primaryClass={cs.LG}
}

@article{vinyals2019grandmaster,
  title={Grandmaster level in StarCraft II using multi-agent reinforcement learning},
  author={Vinyals, Oriol and Babuschkin, Igor and Czarnecki, Wojciech M and Mathieu, Micha{\"e}l and Dudzik, Andrew and Chung, Junyoung and Choi, David H and Powell, Richard and Ewalds, Timo and Georgiev, Petko and others},
  journal={Nature},
  volume={575},
  number={7782},
  pages={350--354},
  year={2019},
  publisher={Nature Publishing Group}
}

@article{cobbe2020phasic,
      title={Phasic Policy Gradient}, 
      author={Karl Cobbe and Jacob Hilton and Oleg Klimov and John Schulman},
      year={2020},
      eprint={2009.04416},
      archivePrefix={arXiv},
      primaryClass={cs.LG}
}

@incollection{pytorch,
    title = {PyTorch: An Imperative Style, High-Performance Deep Learning Library},
    author = {{PyTorch Contributors}},
    booktitle = {Advances in Neural Information Processing Systems 32},
    editor = {H. Wallach and H. Larochelle and A. Beygelzimer and F. d\textquotesingle Alch\'{e}-Buc and E. Fox and R. Garnett},
    pages = {8024--8035},
    year = {2019},
    publisher = {Curran Associates, Inc.},
    url = {http://papers.neurips.cc/paper/9015-pytorch-an-imperative-style-high-performance-deep-learning-library.pdf}
}

@Article{numpy,
 title         = {Array programming with {NumPy}},
 author        = {{NumPy Contributors}},
 year          = {2020},
 month         = sep,
 journal       = {Nature},
 volume        = {585},
 number        = {7825},
 pages         = {357--362},
 doi           = {10.1038/s41586-020-2649-2},
 publisher     = {Springer Science and Business Media {LLC}},
 url           = {https://doi.org/10.1038/s41586-020-2649-2}
}

@software{pandas,
  author       = {{Pandas Contributors}},
  title        = {pandas-dev/pandas: Pandas 1.0.3},
  month        = mar,
  year         = 2020,
  publisher    = {Zenodo},
  version      = {v1.0.3},
  doi          = {10.5281/zenodo.3715232},
  url          = {https://doi.org/10.5281/zenodo.3715232}
}

@ARTICLE{scipy,
  author  = {{SciPy Contributors}},
  title   = {{{SciPy} 1.0: Fundamental Algorithms for Scientific
            Computing in Python}},
  journal = {Nature Methods},
  year    = {2020},
  volume  = {17},
  pages   = {261--272},
  adsurl  = {https://rdcu.be/b08Wh},
  doi     = {10.1038/s41592-019-0686-2},
}

@Article{ipython,
  Author    = {P\'erez, Fernando and Granger, Brian E.},
  Title     = {{IP}ython: a System for Interactive Scientific Computing},
  Journal   = {Computing in Science and Engineering},
  Volume    = {9},
  Number    = {3},
  Pages     = {21--29},
  month     = may,
  year      = 2007,
  url       = "https://ipython.org",
  ISSN      = "1521-9615",
  doi       = {10.1109/MCSE.2007.53},
  publisher = {IEEE Computer Society},
}

@Article{matplotlib,
  Author    = {Hunter, J. D.},
  Title     = {Matplotlib: A 2D graphics environment},
  Journal   = {Computing in Science \& Engineering},
  Volume    = {9},
  Number    = {3},
  Pages     = {90--95},
  publisher = {IEEE COMPUTER SOC},
  doi       = {10.1109/MCSE.2007.55},
  year      = 2007
}

@inproceedings{geotorch,
    title = {Trivializations for gradient-based optimization on manifolds},
    author = {Lezcano-Casado, Mario},
    booktitle={Advances in Neural Information Processing Systems, NeurIPS},
    pages = {9154--9164},
    year = {2019},
}

@software{plotnine,
  author       = {{Plotnine Contributors}},
  title        = {has2k1/plotnine: v0.8.0},
  month        = mar,
  year         = 2021,
  publisher    = {Zenodo},
  version      = {v0.8.0},
  doi          = {10.5281/zenodo.4636791},
  url          = {https://doi.org/10.5281/zenodo.4636791}
}

\appendix

\subsection{AlphaZero Implementation}
\label{alphazero}

While our implementation was heavily influenced by several different open-source AlphaZero implementations \cite{lanctot2020openspiel, wu2020accelerating, tian2019elf, cazenave2020polygames}, our unusual use-case - training small agents on small boards - lead to some unusual design decisions.

\subsubsection{Small networks} The original AlphaZero and its open-source replications used very large residual convnets. ELF OpenGo \cite{tian2019elf}, for example, uses a 256-filter 20-block convolutional network, weighing in at roughly 20m parameters and 2 GF-s for a forward pass on a single sample. In our preliminary work however, we found that on the small boards we work with, far smaller - and faster - networks could make it to perfect play.

In particular, we found that perfect play on a $9 \times 9$ board can be achieved by a fully-connected residual net with two layers of 512 neurons, along with an input and output layer. This net weighs in at 500k parameters and 500 KF-s for a forward pass, a tiny fraction of the cost of the original AlphaZero networks.

\subsubsection{Vectorization} These very-small networks open the way to further speedups. When the neural networks involved in a reinforcement learning problem are large, the time taken to forward- and backward-propagate through the network dominates the run time of the algorithm. As such, it doesn't often make sense to invest effort in speeding up other parts of the implementation. When the neural networks are small however, these other-parts come to the fore.  

In contrast to other AlphaZero implementations, where the environment and tree search are implemented on the CPU, our implementation is wholly-GPU based. Both the rules of Hex and the tree search codes are written in CUDA and carried out on the GPU. This enables us to massively parallelise things, with a typical training setup collecting experience from 32k Hex boards in parallel. 

This is a technique that has been implemented now for a range of environments \cite{dalton2020accelerating, megastep}, but ours is the first application of the technique to board games and to MCTS. 

\subsubsection{Regularized tree search} If AlphaZero's tree search discovers a particularly high-value strategy during exploration, it can take many, many simulations before the high value of that strategy is fully reflected at the root node. This issue was identified in Grill et al. \cite{grill2020montecarlo}, which also shows it can be resolved by solving a simple optimization problem at each node. 

We found that adapting their solution let us use dramatically fewer nodes in our search tree. We did however find that in the small-network regime this work is concerned with, the bisection search proposed by Grill et al. can be a significant factor in the runtime. Fortunately this issue was easily resolved by replacing the bisection search with a Newton search.

We also discovered that the ideal coefficient of exploration, $c_\text{puct}$, was far lower than advertised elsewhere in the literature. This corresponds to a lower emphasis on the prior policy distribution versus the findings of the tree search. Our ideal value was in the region of $1/16$, compared to the $2$ used in prior work. We remain uncertain as to whether this is due to some peculiarity of our setup, or a consequence of our use of regularized tree search. 

\subsubsection{Large batches} It has been observed that many reinforcement learning schemes are substantially faster and more stable when large batch sizes are used \cite{stooke2019accelerated}. 

In typical AlphaZero implementations however, the batch size is typically $\approx 1000$ samples. We are not certain as to why this is the case, but suspect it is a consequence of large size of the value/policy network limiting how much can be held in memory at once. With our particularly-small networks however, this is much relaxed, and so our runs typically use a batch size of 32k samples. This size was arrived at by calculating the gradient noise scale \cite{mccandlish2018empirical}, which is roughly 32k on the largest boards.

\subsubsection{Small buffer} A final discovery was that while many other multi-agent training schemes include large replay buffers \cite{silver2018general}, tournaments \cite{vinyals2019grandmaster} and leagues \cite{openai2019dota} as a way to suppress cyclic patterns during training, we found that none of these were necessary in our implementation. We do not know if this is a consequence of our choice of game, our small board sizes, our small agents, or our large batches, but the outcome is that we could use a very small replay buffer - 2m samples, or 64 steps of our 32k-replica environment. This lead to a dramatic speedup in training, plausibly due to the much lower staleness of the samples ingested by the learner \cite{cobbe2020phasic, openai2019dota}.

\subsubsection{Validity}
In all we have made \emph{many} novel adjustments to AlphaZero in this work, and if we were claim superiority over a more conventional implementation then we would be obliged to present a wide array of baselines and ablations. However, this work's goal is not to present a fast Hex solver. The critical property is simply whether our implementation can achieve perfect play, and by comparison to MoHex we find ourselves suitably convinced of this.

\subsection{Handling Non-Transitivity} 
As discussed in §\ref{background-elo}, Elo ratings are non-transitive. One worry we had was that the compute frontiers observed here might be a product of this non-transitivity and the varying numbers of agents used at different board sizes. To resolve this worry we also tried evaluating every agent directly against a single top-rated agent.

This `top-agent' evaluation has no transitivity issues, but does require matchups between agents of vastly different skill levels. The 2,000 Elo difference between random and perfect play on a $9 \times 9$ board (Fig. \ref{flops_curves}) implies that the random agent will win 1 in 1m games against the perfect agent. This means we would need to play $\gg$10m games to properly resolve the rating of the random agent. 

\begin{figure}[tb]
\centerline{\includegraphics[width=.49\textwidth]{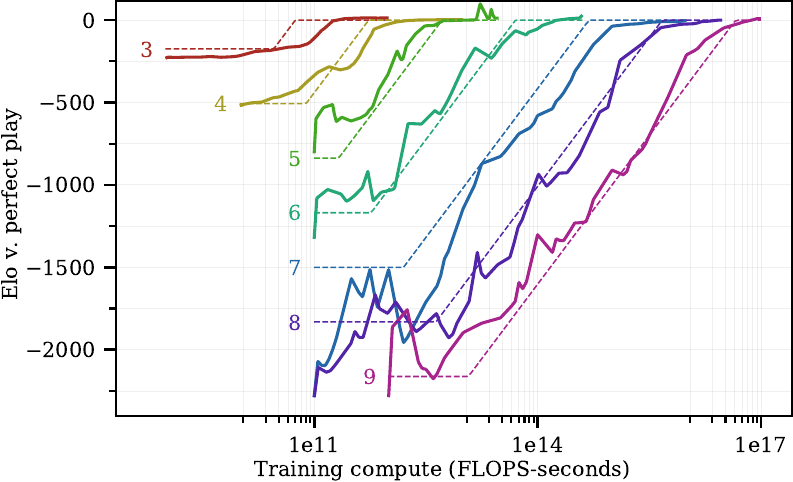}}
\caption{When computed using top-agent evaluation instead, the frontiers are noisier than in league evaluation but display the same form and similar fits.}
\label{direct_frontiers}
\end{figure}

While this is more than we could afford across the 2,800 agents in our stable, we decided to play a limited number of games - 64k - between each agent and a top-rated agent. We found that the frontiers derived using this setup were noisier than the ones generated by playing every agent against every other, but that the pattern was similar in form and fit. The frontiers from this evaluation method can be seen in Fig. \ref{direct_frontiers}, and give the fitted parameters in Table \ref{direct-parameters}. 

\begin{table}[t]
\centering
\caption{Fitted Frontier Parameters (Top-Agent Evaluation)}
\label{direct-parameters}
\begin{tabular}{llll}
\toprule
{} & $m_\text{flops}$ & $m_\text{boardsize}$ &    $c$ \\
\midrule
plateau &                  &                 -330 &    820 \\
incline &              600 &                 -590 &  -4700 \\
\bottomrule
\end{tabular}
\end{table}

In all, we are convinced that the compute frontiers observed are not due to non-transitivity. 

\subsection{Alternate curve models}
\label{curve}

We experimented with several functional forms for the compute frontiers. 

Linear models were our first choice, but the notably non-linear behaviour at the top and bottom of each curve damaged the estimates of the slope of each frontier and the compute required for perfect play.

Sigmoid models meanwhile were much better fits, and their smoothness is arguably a better fit for the phenomena in question. However, that same smoothness makes interpreting their parameters much harder. 

The change-point model used in the main text is as good of a fit (in terms of MSE) as the sigmoid model, but its parameters are much easier to interpret. 

\subsection{Software}
This work depended principally on the libraries pytorch \cite{pytorch}, geotorch \cite{geotorch}, numpy \cite{numpy}, scipy \cite{scipy}, pandas \cite{pandas}, ipython \cite{ipython}, matplotlib \cite{matplotlib} and plotnine \cite{plotnine}.

\end{document}